\documentclass[twoside,leqno,twocolumn]{article}

\usepackage[letterpaper]{geometry}

\usepackage{ltexpprt}

\usepackage{booktabs} % For formal tables
\usepackage{amsmath}
\usepackage{algorithm}
\usepackage{algorithmicx}
\usepackage{algpseudocode}
\usepackage{hyperref}
\usepackage{xcolor}

\usepackage{graphicx}

\graphicspath{{figures/}}
\DeclareGraphicsExtensions{.pdf,.jpeg,.png}
\usepackage[tight,footnotesize]{subfigure}
\usepackage{comment}
\makeatletter
\def\BState{\State\hskip-\ALG@thistlm}
\makeatother
\algdef{SE}[DOWHILE]{Do}{doWhile}{\algorithmicdo}[1]{\algorithmicwhile\ #1}%
\algrenewcommand\algorithmicindent{1.0em}

\usepackage{mdwmath}
\usepackage{mdwtab}
\usepackage{eqparbox}
\usepackage{multirow}
\usepackage[tight,footnotesize]{subfigure}
\usepackage{caption}
\usepackage[tight]{subfigure}
\usepackage[]{xcolor}

\newif\ifdraft
\draftfalse
\ifdraft
\newcommand{\dipendra}[1]{\textcolor{Red}{[Dipendra: #1]}\xspace}
\else
\newcommand{\dipendra}[1]{\xspace}
\fi

\begin{document}
%\title{Phase Mapping using Fuzzy Binary Representations of X-ray Diffraction Patterns}
\title{An Incremental Phase Mapping Approach for X-ray Diffraction Patterns using Binary Peak Representations\thanks{Accepted and presented at the International Workshop on Domain-Driven Data Mining (DDDM) as a part of the SIAM International Conference on Data Mining (SDM'2021).}}

\author{Dipendra Jha\thanks{Department of Electrical and Computer Engineering, Northwestern University, Evanston, IL, United States} \and
K.V.L.V. Narayanachari\thanks{Department of Materials Science and Engineering, Northwestern University, Evanston, IL, United States} \and
Ruifeng Zhang\footnotemark[2]\and
Justin Liao\thanks{Department of Computer Science, Brown University, Providence, RI, Unites States} \and
Denis T. Keane\thanks{DND-CAT Synchrotron Research Center, Northwestern University, Evanston, IL, United States},
Wei-keng Liao\footnotemark[1] \and
Alok Choudhary\footnotemark[1] \and
Yip-Wah Chung\footnotemark[2] \and
Michael Bedzyk\footnotemark[2] \and
Ankit Agrawal\footnotemark[1] \thanks{Corresponding author, email: ankit-agrawal@northwestern.edu}
}
% \email{dipendra.jha@eecs.northwestern.edu, achari.kondapalli@northwestern.edu, \\ ruifengzhang2019@u.northwestern.edu, jrex.liao@gmail.com \\ dtkeane@northwestern.edu, \\ wkliao@eecs.northwestern.edu, choudhar@eecs.northwestern.edu, \\
% ywchung@northwestern.edu, bedzyk@northwestern.edu, ankitag@eecs.northwestern.edu}

\date{}

\maketitle

% % Default Copyright Statement
% \fancyfoot[R]{\scriptsize{Copyright \textcopyright\ 2021 by SIAM\\
% Unauthorized reproduction of this article is prohibited}}

\begin{abstract}
%X-ray diffraction (XRD) is a widely used experimental technique that provides the atomic-level structural information of material samples which determines their composition-structure-property relationships for new materials design and discovery. 
Despite the huge advancement in knowledge discovery and data mining techniques, the X-ray diffraction (XRD) analysis process has mostly remained untouched and still involves manual investigation, comparison, and verification. Due to the large volume of XRD samples from high-throughput XRD experiments, it has become impossible for domain scientists to process them manually. Recently, they have started leveraging standard clustering techniques, 
to reduce the XRD pattern representations requiring manual efforts for labeling and verification. Nevertheless, these standard clustering techniques do not handle problem-specific aspects such as peak shifting, adjacent peaks, background noise, and mixed phases; hence, resulting in incorrect composition-phase diagrams that complicate further steps. Here, we leverage data mining techniques along with domain expertise to handle these issues. In this paper, we introduce an incremental phase mapping approach based on binary peak representations using a new threshold based fuzzy dissimilarity measure. The proposed approach first applies an incremental phase computation algorithm on discrete binary peak representation of XRD samples, followed by hierarchical clustering or manual merging of similar pure phases to obtain the final composition-phase diagram. We evaluate our method on the composition space of two ternary alloy systems- Co-Ni-Ta and Co-Ti-Ta. Our results are verified by domain scientists and closely resembles the manually computed ground-truth composition-phase diagrams. The proposed approach takes us closer towards achieving the goal of complete end-to-end automated XRD analysis.

\end{abstract}

% \begin{Keywords}
% X-ray Diffraction, Phase Clustering, Unsupervised Learning, Composition-Phase Diagram, Fuzzy Representation
% \end{IEEEkeywords}

\section{Introduction}
X-ray diffraction (XRD) combined with X-ray fluorescence (XRF) is a well-known composition spread experimental technique used for the mapping the composition-structure-property relationships for high-throughput materials discovery~\cite{woolfson1997introduction, klug1974x,moore1989x,bish1989modern,
cullity1978elements}.
XRD analysis provides scientists with the atomic-level information for structural characterization of materials.
The diffraction peaks in a XRD pattern reflect the atomic arrangement within the material crystal
structure, which not only determines the properties of materials but also helps in assessing the phases in the material structure for new materials design and discovery~\cite{chung1999automated}.

Generally human experts analyze the XRD patterns by using their domain knowledge to examine the peak characteristics such as peak spread and location, which are then correlated with the sample composition and known phases to identify the phase in the measured sample.
Since the XRD patterns are often noisy,
% due to a collection of issues including background radiation, detector noise, and sample-detector configuration,
%which may result into significantly varying background noise from sample to sample
%~\cite{mos2018x}.
% The presence of highly irregular background makes the peak search process more complicated.
the first step in XRD analysis is to remove the background noise by fitting a background curve~\cite{mos2018x, savitzky1964smoothing, proctor1980smoothing,
dinnebier2003fwhm, bruckner2000estimation, kajfosz1987nonpolynomial}.
This is followed by indexing the peak positions against existing reference databases, and correlating with the sample composition to identify the phases in the measured sample using software tools, which often requires verification by a domain expert.
The phase of the material at a given composition represents the specific chemistry and atomic arrangement in the crystal structure of the material sample \cite{cullity1978elements}.
Current high-throughput XRD experiments produce thousands of samples in each composition-spread experiment, making the manual attribution of sample phase labeling a very arduous task.

Recently domain scientists have begun to leverage machine learning techniques to obtain the potential composition-phase diagram~\cite{tatlier2011artificial, gilmore2004high, bunn2016semi}. 
Standard clustering techniques such as k-means clustering and hierarchical clustering, help in reducing the number of samples that require manual phase labeling and verification.
% Once a subset of samples are manually labelled and validated by domain experts, supervised learning techniques have been used to label the rest of the samples~\cite{bunn2016semi}.
% Recently, Park et al.~\cite{park2017classification} used a CNN to classify 1D XRD patterns into space group, extinction-group and crystal-system classes.
% Since the experimental XRD patterns with labels are impossible to obtain for a large enough dataset to train a CNN, they used 150000 XRD patterns computed from the structure solutions of entries in the Inorganic Crystal Structure Database~\cite{icsd}.
% Hence, the machine learning technique used by domain scientists have been mainly limited to unsupervised and semi-supervised techniques due to lack of labeled datasets, since manual labeling of large volume of XRD samples is a very arduous task.
Nevertheless, the current practice of using standard clustering techniques for XRD analysis do not handle the problem-specific aspects of XRD such as peak shifting, adjacent peaks, background noise and mixed phases.
For instance, the existing distance metrics used by domain scientists treat the intensity values along the Q-value axis as independent features without considering the adjacency of peaks.
Although some dissimilarity measures such as dynamic time warping (DTW), and the earth mover's distance, are specifically developed for feature preservation with resilience to phase shifting, these metrics perform worse than the geometry based metrics on 1D XRD patterns~\cite{iwasaki2017comparison}.
Also, a composition-phase diagram is generally composed of multiple pure phases and their mixed phases.%; the mixed phase being composed of multiple constituent pure phases.
The standard clustering used by domain scientists is a hard clustering approach which treats the mixed phase as a separate individual cluster from its constituent pure phases.%, leading to incorrect phase clusters.
Hence, the standard clustering techniques used for XRD analysis often results in incorrect potential composition-phase diagram, which further complicates the manual labeling and verification.
% Furthermore, these standard clustering techniques require a lot of manual parameter tuning in large parameter spaces for proper phase clustering.
% For example, there exist multiple distance metrics for the cluster analysis of XRD data- geometry based measures such as $L_1$ norm ('Manhattan'), $L_2$ norm ('Euclidean'), cosine metric, and statistics-based dissimilarity measures such as Pearson correlation coefficient and Spearman rank correlation coefficient.
Here, we leverage data mining techniques with domain expertise to handle these issues and automate the high-throughput XRD analysis process.% of high-throughput XRD experiments.

% The current practice of directly using existing clustering algorithms has a key issue with the use of distance metrics and 
% Another main issue with current technique is that it involves a lot of manual parameter tuning for proper phase clustering output.
% Also, the composition-phase diagrams are generally composed of multiple pure phases and mixed phases;
% Although a mixed phase being a composition of multiple pure phases,
% the current approach of hard clustering treats each of these pure and mixed phases as separate clusters which leads to improper phase clusters.
% % Since the clustering involves comparing against each sample in the dataset using value of peak intensity at each Q-value, they are computationally expensive.
% Here, our goal is to address this issue by developing a phase mapping technique based on binary peaks using a fuzzy dissimilarity measure which considers the adjacency along the peak location (Q-axis) and could automatically consider the potential pure and mixed phases.

In this paper, we introduce an incremental phase mapping approach based on binary peak representations using a new threshold based fuzzy dissimilarity measure for high-throughput XRD analysis.
% The peak locations are the most critical information present in a XRD sample, without any well defined adjacency threshold.
The proposed method begins by converting the XRD sample into discrete binary peak representations where the presence and absence of peaks are represented using 1 and 0.
Next, we develop an incremental (pure) phase computation algorithm based on discrete binary peak representations using a new threshold based fuzzy dissimilarity measure for 1D XRD patterns.
% The mixed phases for a particular peak count are computed by taking different combinations of pure phases with lower peak counts.
This step outputs the initial composition-phase diagram along with the discrete binary pure phase representations; these pure phase representations are simple to analyze due to their discrete binary representations and can be used by domain scientists for further analysis and validation.
% Finally, we can either apply hierarchical clustering on the initial pure phase representation or manually merge the similar ones to obtain the final composition-phase diagram.
We evaluate the proposed approach on XRD samples from high-throughput composition-spread experiments of two ternary alloy systems- Co-Ni-Ta and Co-Ti-Ta.
The results using proposed approach are verified by domain scientists and agrees with the manually computed ground truth composition-phase diagram. 

\section{Background}
X-ray diffraction combined with X-ray fluorescence is a well-known atomic scale probing technique to determine the crystal structure of materials for mapping the composition-structure-property relationships for high-throughput materials discovery~\cite{klug1974x, moore1989x, bish1989modern, cullity1978elements,woolfson1997introduction}.
During an XRD experiment, the atomic arrangement of the crystal structure in the materials diffracts the beam of incident X-rays into many specific directions; using the diffraction angles, one can construct a unit cell to represent the atomic locations in the crystal by measuring the diffraction angles and X-ray intensities of the diffraction pattern.

\begin{figure*}[!htb]
\centering
\includegraphics[width = 0.7\linewidth, keepaspectratio=true]{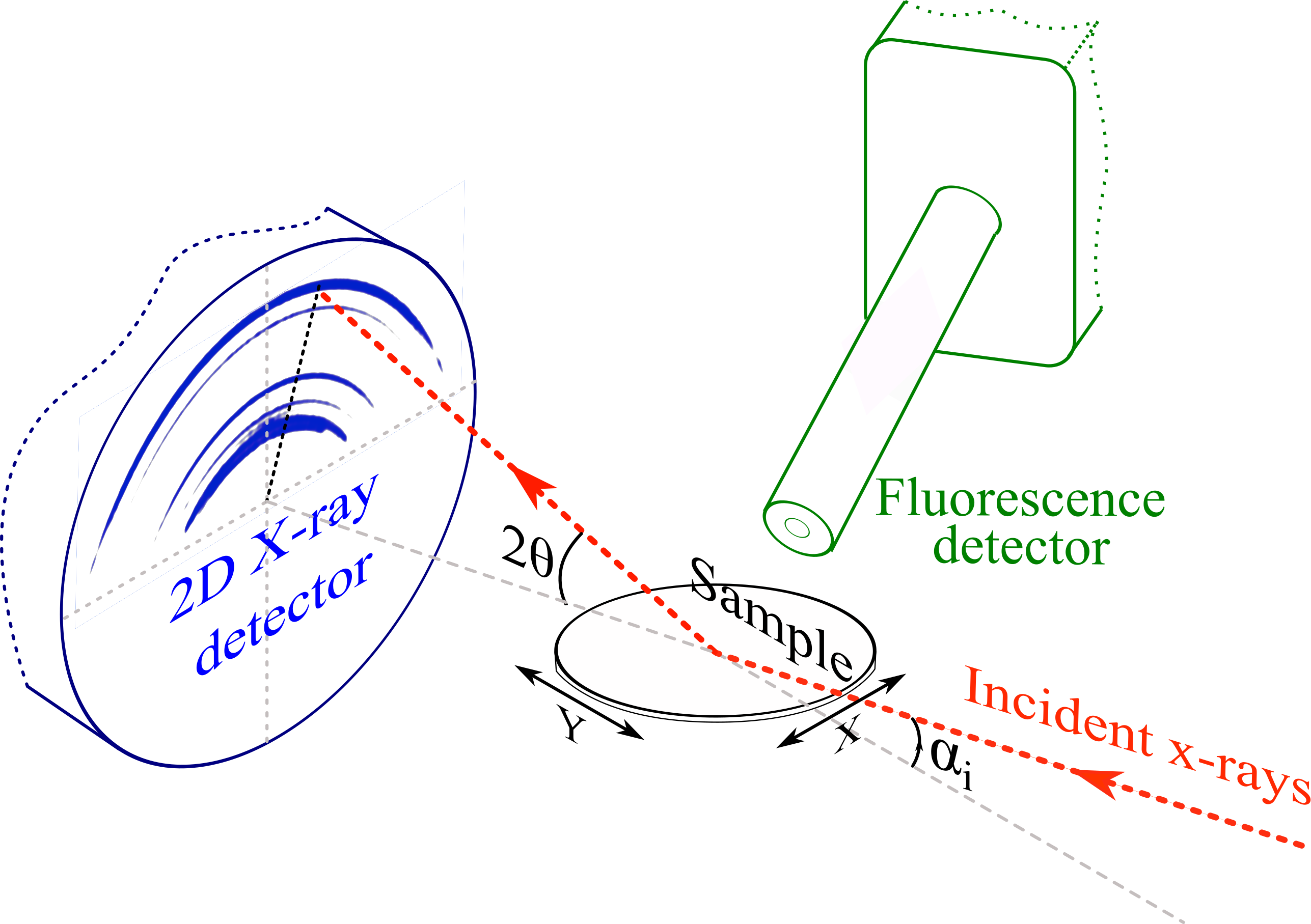}

\caption{A schematic representation of the high-throughput X-ray diffraction experimental setup at beam line 5MBC at APS in Argonne National Lab. The sample stage moves laterally to probe various locations on the sample (100 mm diameter) with a pitch of 2 mm in both X and Y directions. The projected beam size on the sample surface is 500 $\mu$ m $\times$ 500 $\mu$ m. The incidence angle $\alpha _{i}$ is fixed at 10$^\circ$.}
\label{fig:setup}
\end{figure*}

An XRD diffraction pattern is an intensity plot of the X-rays scattered at different angles by the material sample, as measured by a 2D detector, with value at each pixel representing the count of incident X-ray photons.
The X-ray diffraction pattern from a given material sample is composed of multiple sharp spots that represent the periodic atomic arrangement in the crystal of the material; these sharp spots are known as Bragg diffraction peaks.
By measuring the position and intensities of the Bragg diffraction peaks, one can determine the phase of the material- the specific chemistry and atomic arrangement in the crystal structure of the material sample \cite{cullity1978elements}.
Materials with same composition can have multiple crystal structures, and hence phases. For instance, diamond, charcoal and graphite are different phases of carbon; they are chemically identical but follow different atomic arrangement in their crystal structures.
A phase map contains the different constituent phases representing different crystal lattice structures for varying materials composition for a given material system.
The clustered data can be represented on a ternary phase diagram as a function of atomic fractions. The phase clustering result can also be visualized by plotting a circular plot that represent the phases as a function of X-Y coordinates of the wafer used during XRD experiment.
More details about X-ray powder diffraction can be found in Scott et al.~\cite{prismmit,cullity1978elements}. 

\section{Method}
The proposed method first converts the XRD samples into discrete binary peak representations. Next, we apply an incremental phase computation algorithm on discrete binary peak representation of XRD samples, followed by hierarchical clustering or manual merging of similar initial pure phases if required, to obtain the final composition-phase diagram.

\subsection{Binary Peak Representation}

The conversion of XRD samples into discrete binary peak representations is accomplished using three steps.
First, we fit a high degree polynomial on top of the 1D XRD pattern inside a fixed window to eliminate the intensity spikes and irregular background.
Second, we compute the baseline of lower degree on the previous output to obtain the background noise and subtract it to get the processed XRD pattern.
Final step is to detect the presence of peaks in the processed XRD pattern by using a threshold value for intensity count and a fixed window size representing the bin.
The peaks are searched by taking the first order difference of the processed XRD pattern.
Once we obtain the peak positions, we compute the binary peak representation for the given sample by assigning 1 and 0 respectively to represent the presence and absence of a peak in a given window.

% First, a polynomial curve of degree $deg$ is fit on the XRD with a window of size $window$. 
% This removes the intensity spikes and irregular background from the XRD pattern.
% Next, we compute a baseline of degree $base\_line$ on the previous output signal and subtract the baseline from previous output to remove the background noise and obtain the $proc\_signal$.
% The last step in binarization is to detect the presence of peaks using a threshold value of $thresh$ and a window of size $min\_dist$ from the processed signal $proc\_signal$.
% The peaks are searched by taking the first order difference of the signal $proc\_signal$.
% After getting the peak positions, we compute the discrete binary representations for the given sample by assigning 1 to the window which contains a peak and 0s elsewhere.

% \begin{algorithm}
% \caption{ConvertBinary}
% \label{alg:binary}
% \begin{algorithmic}[1]
% \Statex \textbf{Input:} $sample, window, thresh, min\_dist,$
% \Statex $deg, base\_deg$
% \Statex \textbf{Output:} {$binary\_pattern$}
% \State $signal \gets GetIntensities(sample)$
% \State $poly\_signal \gets FitPolynomial(signal, window, deg)$
% \State $base\_signal \gets FitPolynomial(poly\_signal, base\_deg)$
% \State $proc\_signal \gets poly\_signal - base\_signal$
% \State $peak\_indexes \gets FindIndexes(proc\_signal, thresh, min\_dist)$
% \State $binary\_pattern \gets GenerateBinarySignal(sample, peak\_indexes)$
% \end{algorithmic}
% \end{algorithm}

\subsection{Incremental Phase Computation}

The proposed phase computation algorithm is based on the critical and valid assumption that all the samples belonging to the same phase (having same underlying crystal structure) have same peak counts with similar peak locations. Note that this assumption puts critical emphasis on the choice of threshold value for intensity count and the fixed window size for detecting peaks during conversion of XRD patterns to their corresponding binary peak representations.
The algorithm starts by first organizing the input data into groups with same peak counts.
Note that the input for this step is the binary peak representation of the XRD samples in discrete form.
The phase computation is performed in an incremental manner on each group using Algorithm~\ref{alg:cphelper}; phase computation algorithm is applied on each group in increasing order of peak count.

\begin{algorithm}[!tb]
\caption{PhaseComputation}
\label{alg:cphelper}
\begin{algorithmic}[1]
\State { Input:} $data, pc, th, pp, mm$.
\State { Output:} $pp, mm$.
% \If{$len(pure\_rep) \gets 0$}
% \State $start\gets 1$
% \State $pure\_res \gets data[0]$
% \State $spi \gets 0$
% %\State $pure\_counter \gets list([1])$
% \State $membership \gets list([0,-1,-1])$
% \Else
% \State $start\gets0$
% \State $spi \gets len(pure\_reps)$
% \EndIf
\State $mp \gets $ComputeMixedPhases$(pp, pc, th)$.
% \State $mixed\_phases, memb\_mp \gets GetMixedPhases($
% \Statex $pure\_reps, num\_peaks, ploc\_thresh)$
%\State $spi \gets len(pure_reps)-1$
\For{$sample \in data$}
%\State $candidate \gets data[i]$
\State $new \gets True$.
\For{$p \in pp$}
\State $eq \gets $FuzzyEquals$(sample, p, th)$.
\If{$eq$}
\State $new \gets False$.
%\State $pure\_counter[j] \gets pure\_counter[j]+1$
\State Update $mm$ using $p$.
\State break.
\EndIf
\EndFor
%\If{not $new$}
%\State break
%\EndIf
\If{$new = False$}
\State Continue.
\EndIf
\For{$p \in mp$}
\State $eq \gets $FuzzyEquals$(sample, p, th)$.
\If{$eq$}
\State $new \gets False$.
% \For{$n \in range(3)$}
% \If{$belongs\_to[j,n] > 0$}
% \State Add $1$ to $pure\_counter[memb\_mp[j,n]]$
% \EndIf
% \EndFor
\State Update $mm$ using $p$.
\State break.
\EndIf
\EndFor
\If{$new = True$}
\State Append $sample$ to $pp$.
\State Update $mm$ using $sample$ as new phase.
%\State Append $1$ to $pure\_counter$
\EndIf
\EndFor
\For{$p \in pp$}
\If{$p$ contains $pc$ peaks}
\State Update $p$ using average representation of patterns in $data$ belonging to pure phase $p$.
\EndIf
\EndFor

\end{algorithmic}
\end{algorithm}

%\subsubsection{Fuzzy distance metric}

\begin{algorithm}[tbh]
\caption{FuzzyEquals}
\label{alg:fuzzyequals}
\begin{algorithmic}[1]
\State { Input:} $s1, s2, th$.
\State $l1 \gets GetPeakLocations(s1)$.
\State $l2 \gets GetPeakLocations(s2)$.
\If{$l1$ and $l2$ differ in peak counts}
\State \Return $False$.
\EndIf
\For{$i \in l1$}
\If{Distance between $i$ and nearest peak in $loc2 > th$}
\State \Return $False$.
\EndIf
\EndFor
\For{$i \in l2$}
\If{Distance between $i$ and nearest peak in $loc1 > th$}
\State \Return $False$.
\EndIf
\EndFor
\State \Return True
\end{algorithmic}
\end{algorithm}

Algorithm~\ref{alg:cphelper} takes a group of XRD samples ($data$) along with the peak count for current group ($pc$), adjacency threshold for peak locations ($th$), existing pure phases ($pp$) as the input parameters and returns all the pure phase representations including the new ones with current peak count ($pp$) and phase membership for all input samples ($mm$) as the output.
This algorithm starts by first computing the mixed phases with the current peak count ($pc$) using different combinations of existing pure phases in $pp$.
Since a mixed phase is a combination of multiple pure phases, a sample with a mixed phase contains peaks at all of its corresponding constituent pure phase peak locations.
Note that the current practice using standard clustering do not handle samples with mixed phases correctly since it assigns a mixed phase to its own separate cluster, rather than assigning a mixed phase to its multiple constituent pure phase clusters.
Next, each sample is compared with the already computed pure phases and mixed phases with $pc$ peaks using a threshold based fuzzy dissimilarity measure in Algorithm~\ref{alg:fuzzyequals}.
If a match is found, the sample is assigned to the matching phase by appending its membership to $mm$.
If no match is found, the current sample is assigned to a new pure phase (and appended to $pp$) and $mm$ is updated accordingly.
After the phase computation is done for all the samples in the input $data$, the pure phase representations for the current group is updated using the average representation of all the samples assigned to that particular pure phase.
The updated pure phase representations are used to compute the mixed phases for the next groups of samples with higher peak counts.
Since the phase computation is done for one group of samples with same peak count ($pc$) at a time, all the samples belonging to same cluster have the same peak counts, hence, holding the assumption at the beginning of our algorithm.

Algorithm~\ref{alg:fuzzyequals} takes the two samples for comparison as the inputs along with the adjacency threshold for peak locations ($th$). It works by first checking that the two input samples have same peak counts and then checks whether the peak locations in the two input samples obey the adjacency threshold requirement of $th$.
The two input samples have to meet all these criteria for them to match (belong to same phase cluster).
Hence, all the samples belonging to same phase cluster contain equal number of peaks such that their corresponding matching peaks lie within a fixed peak window.
Since the comparison between the two input samples in done by looking at their fuzzy peak locations rather than the exact positions, we refer to this dissimilarity metric as $FuzzyEquals$.
Note that XRD patterns contain a lot of irregular and regular background noise, which may not be removed using the polynomial fitting and baseline subtraction during conversion to binary peak representation. To handle such samples with peaks resulting from intensity spikes from background noise due to different factors, we use a threshold count for outlier ($ot$). If an initial pure phase from the above phase computation step does not have sufficient samples, we treat such pure phases as outliers and remove them from further consideration. Hence, each of the initial pure phases from the phase computation step has at least $ot$ samples.

\section{Empirical Evaluation}
% We evaluate the proposed phase mapping approach using two high-throughput XRD datasets in this section and present our results and analysis, along with comparison against the current practice of using standard clustering algorithms. All the algorithms and analysis are implemented and performed using Python.

\subsection{Datasets}

The XRD datasets used in this study are collected from two different high-throughput composition spread using concurrent X-ray diffraction (XRD) and X-ray fluorescence (XRD) experiment for the composition space of Cobalt, Nickel and Tantalum (Co-Ni-Ta) and the composition space of Cobalt, Titanium and Tantalum (Co-Ti-Ta).  
The material libraries were  deposited using co-sputter deposition on a 100 mm diameter single crystal sapphire wafer, and further treated at high temperature in vacuum furnace. The data acquisition experiments were carried out using a customized setup at beam-line 5BMC of Advanced Photon Source (APS) in Argonne National Lab. The collected XRD data is used for XRD analysis in this paper; the XRF data is used for computing the atomic ratios.

The 2D XRD patterns collected from the X-ray detector were converted to 1D data by circular averaging the counts to reduce the effect of texturing on the XRD data. The 1D data is further processed to account for the incident beam brightness change across the samples and the background subtracted using 2nd order polynomial fit to the background. Then the angular position  2$\theta$ of the diffracted X-ray peak is converted to Q-values. The Q-values are independent of X-ray energy and directly related to the inverse d-spacing on the planes diffracting the X-rays.
The Co-Ni-Ta dataset contains 1,791 XRD samples with intensities at 499 Q-values in the range of [1.0, 4.2] from the X-ray diffraction experiment.
Co-Ti-Ta samples contain the values of intensities at 1,450 Q-values in the range of [1.0, 4.3] for 1,533 samples from X-ray diffraction experiment.
The dataset also contains the composition of each sample along with the X-Y position on the wafer used during experimentation; the composition is used to plot the ternary plot and the X-Y position is used for plotting the circular plot for XRD analysis.

\subsection{Conversion to Binary Peak Representation}

\begin{figure}
\centering
%\subfigure[]{
\includegraphics[width = 0.99\linewidth, keepaspectratio=true]{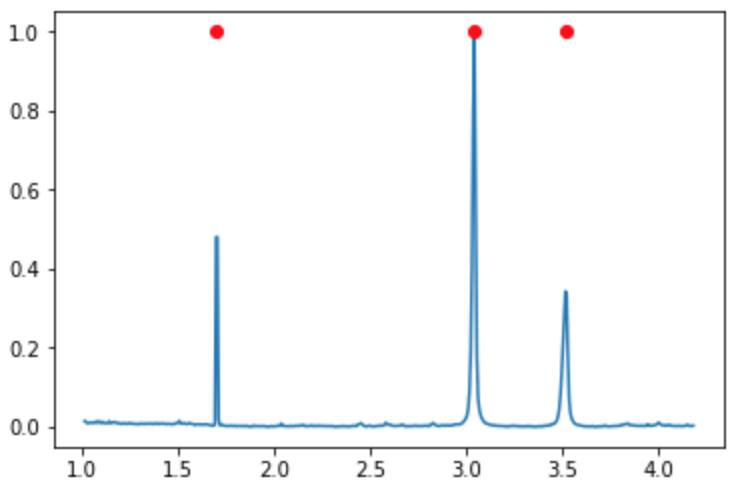}
%}
\caption{XRD sample with it's binary peak representation. The binary peak representation of a 1D XRD pattern contains 1s and 0s to represent the presence and absence of peaks.}

\label{fig:binary_res}

\end{figure}

The proposed phase computing algorithm takes the binary peak representation of XRD patterns as input.
Figure~\ref{fig:binary_res} shows a sample binary peak representation for a given XRD pattern. 
Note that peak locations are the most critical information present in the XRD pattern, hence, we focused on the peak locations to compute the phase of a given material sample in this paper. In our evaluation, we fit a polynomial of degree 5 on the XRD pattern to remove the intensity spikes and irregular background.
This is followed by computing a baseline of degree 1 to obtain the background noise, which is subtracted from the previous output to get the processed XRD pattern.
Final step is to detect the presence of peaks by using a threshold value for intensity count using a fixed window.
The values for threshold and window size have a critical impact on the performance of the proposed approach.
A lower value of threshold results a large number of binary peaks, which may even include the minute peaks from background noise, while higher values of threshold parameter can lead to loss of significant peaks.
Similarly, window size governs the adjacency of peaks during the conversion; lower value of window size results into too many binary values, and higher value for window size gives small number of binary values for the peak representation. The lower value for window size increases the number of bins constituting each peak, hence, the computation cost for next steps.
Within a given window, the peak with highest intensity count is considered to check the threshold criteria.
After thorough experiments for value of threshold and window size, we decided to use 200 windows for Co-Ni-Ta and 1,450 windows for Co-Ti-Ta datasets respectively; hence, the XRD samples in Co-Ni-Ta are represented using 200 binary values and the ones in Co-Ti-Ta are represented using 1,450 binary values denoting the presence and absence of peaks using 1 and 0 values.
The XRD sample in Figure~\ref{fig:binary_res} contains three peak, its binary peak representation contains 1 at the peak position and 0s everywhere else.

\subsection{Phase Mapping Results}

The incremental phase computation algorithm (\ref{alg:cphelper}) takes a group of samples ($data$) along with the adjacency threshold for peak locations ($th$) in the inputs and uses a threshold count for outlier ($ot$) to compute the initial pure phases.
For our evaluation, we experimented with several combinations for the values of these two parameters for the initial phase computation.
The adjacency threshold for peak locations ($th$) decides whether to consider two peak locations during comparison as same or distinct.
A lower value for $th$ results into a large number of initial pure phases; a larger value of $th$ leads to grouping of multiple phases with nearby adjacent peaks into a single phase and hence, overall lower number of probably incorrect pure phases.
The threshold count for outlier ($ot$) is used to decide whether to consider the computed initial pure phase as an outlier or not. If a pure phase does not contain at least $ot$ samples, it is considered an outlier and not removed from further consideration.
Lower values for $ot$ results into a lot of initial pure phases, while higher values lead to elimination of large number of initial pure phases, and their corresponding samples on clustering result output represented by the ternary plot and circular plot.

\begin{figure*}[!t]
\centering
\subfigure[Co-Ni-Ta using $th=1, ot=12$]{
\includegraphics[width = 0.99\linewidth, keepaspectratio=true]{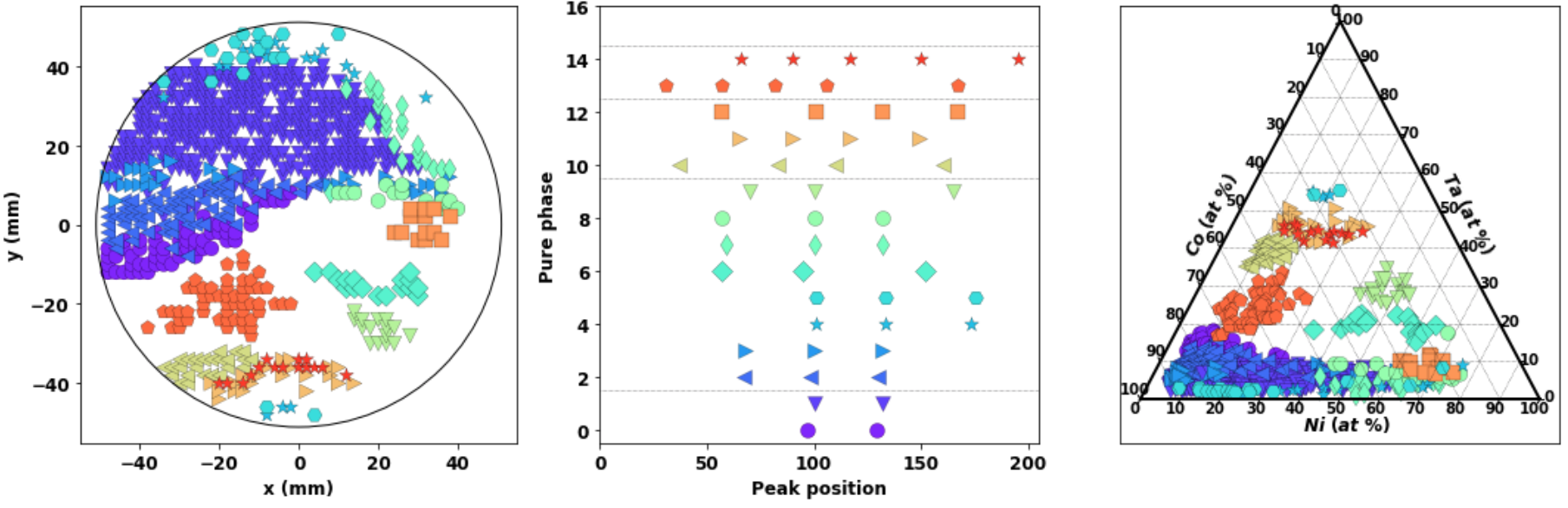}
}
% \subfigure[Co-Ni-Ta using $th=2, ot=13$]{
% \includegraphics[width = 0.99\linewidth, keepaspectratio=true]{CoNiTa-th02-to13.png}
% }
\subfigure[Co-Ni-Ta using $th=4, ot=14$]{
\includegraphics[width = 0.99\linewidth, keepaspectratio=true]{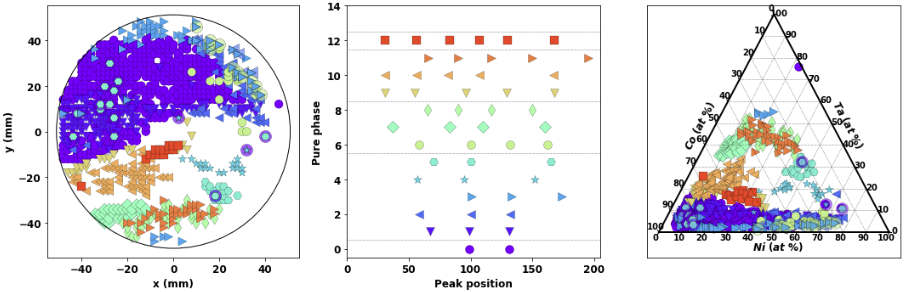}
}
% \caption{Initial phase computation outputs for Co-Ni-Ta composition space using different parameter settings for adjacency threshold for peak locations ($th$) and threshold count for outlier ($ot$). The left subplots show the circular plots representing the phase clustering on wafer using X-Y coordinates, the right subplots show the ternary plot representing the composition-phase diagram (clustering with respect to sample composition), and the middle subplots represent the computed pure phases using binary peak representations to facilitate visualization and further analysis in the next steps by domain experts.}
% \label{fig:results}
% \end{figure}

% \begin{figure}
% \centering
\subfigure[Co-Ti-Ta using $th=20, ot=12$]{
\includegraphics[width = 0.99\linewidth, keepaspectratio=true]{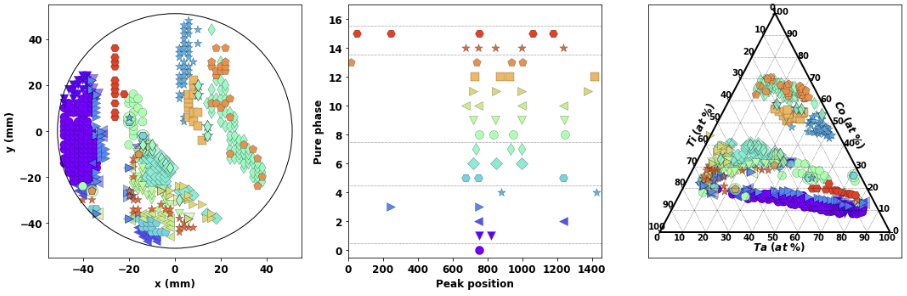}
}
\caption{Initial phase computation outputs using different thresholds parameter settings for adjacency threshold for peak locations ($th$) and threshold count for outlier ($ot$). The left subplot show the circular plot representing the phase clustering on wafer using X-Y coordinates, the right subplot shows the ternary plot representing the composition-phase diagram (clustering with respect to sample composition), and the middle subplot represents the computed pure phases using binary peak representations to facilitate visualization and further analysis for merging similar phases in next step by domain experts.}
\label{fig:results}
\end{figure*}

For Co-Ni-Ta XRD samples, we explored the values for $th$ in the range of [1,30] and for $ot$ in the range of [2,30].
Figure~\ref{fig:results}(a) and (b) show the initial phase computation outputs obtained using different values of $th$ and $ot$.
The pure phase representations shown in the middle subplots can be leveraged to analyze the impact of these parameters in depth.
For instance, there are 8, 6 and 5 initial pure phases with 3 peaks for the $th$ values of 1, 2 and 4, respectively.
In Figure~\ref{fig:results}(a) middle subplot, although 
%the (first and second), (third and fourth), and (sixth and seventh) pure phases with three peaks (from bottom) 
some of the initial pure phases, such as the first and second pure phase with three peaks (counted from bottom)
have similar peak locations, they are considered as two different pure phases due to low value of adjacency threshold for peak locations ($th=1$).
This is also true for the third and fourth pure phases, as well as the sixth and seventh initial pure phases.
We can observe the impact of increasing the value of $th$ if we look at the initial pure phases with two peaks.
As the value of $th$ increases from 1 %and 2 
to 4, the two initial pure phases in the first two cases are merge into a single initial pure phase in the second case (Figure~\ref{fig:results}(b)).

Since Co-Ti-Ta XRD samples contain more fine-grained peaks, we experimented with $th$ values in the range of [2,60] and $ot$ values in the range of [2,30] for this dataset.
Figure~\ref{fig:results}(c) illustrates the initial phase computation output for a particular set of parameter values ($th = 20 $ and $ot = 12$)  for the Co-Ti-Ta composition space.
There exist 16 initial pure phases, almost all of them are significantly distinct from each other as we can visualize in the middle subplot.
The binary representation of the initial pure phases in the middle subplot provides significant insights into the initial phase clustering results, and is extremely helpful to the domain scientists in assisting with the further analysis in conjunction with the corresponding ternary plot and circular plot.
For instance, we observe that the sixth initial pure phase with four peaks contains an extra peak in the far left in addition to all the peaks present in the third initial pure phase with three peaks (from bottom).
From the ternary and circular plot, we find that these two potential phases are overlapping with each other, which demonstrates that they indeed represent the same pure phase; the extra peak in the sixth initial phase with four peaks might have resulted due to the low threshold value for intensity count used during conversion to binary peak representation of the XRD patterns. 
In this way, the domain experts can leverage the initial pure phase representations (presented in the middle subplots) and correlate them with the X-Y position of their corresponding samples (left subplots) and/or their compositions in the ternary plot (right subplots) to decide the actual pure phases and phase regions to obtain the final composition-phase diagram for a given high-throughput composition spread experiment.

The proposed algorithm also automatically handles the mixed phases. 
The mixed phases are combination of multiple pure phases, their XRD patterns contain peaks from more than one pure phases. 
During the computation of initial pure phases, we also compute a set of possible mixed phases using existing initial pure phases for a given number of peak count, and perform comparison against them.
Samples with mixed phases are represented using multiple markers in both circular plots and ternary plots; to make them more clear we use a varying marker size for different constituent pure phases with thicker line boundary used in plotting (Figure~\ref{fig:results}).
We can observe a lot of mixed phases in Figure~\ref{fig:results}(c), where there is a large overlap of different clusters.
Although the proposed algorithm provides a means to automatically handle the mixed phases, still we find that it can miss a lot of mixed phases due to inappropriate choice of the adjacency threshold ($th$) used during initial phase computation.
We can solve this by analyzing the binary peak representation of the initial pure phase representations provided in the middle subplots.
As we can observe in Figure~\ref{fig:results}(a), the third initial pure phase with four peaks appears to be a mixed phase of the pure phases from the sets of (third and fourth) and either (first or second) or (sixth or seventh) initial pure phase with three peaks (from bottom). 
This can be verified by looking at location of the corresponding phases in the the circular plot in the left column. 
Hence, it is a mixed phase. 
Similarly, domain scientists can analyze other pure phase representations to determine whether they are pure phases or mixed phases.

\begin{figure*}[!t]
\centering
%\subfigure[]{
\includegraphics[width = 0.99\linewidth, keepaspectratio=true]{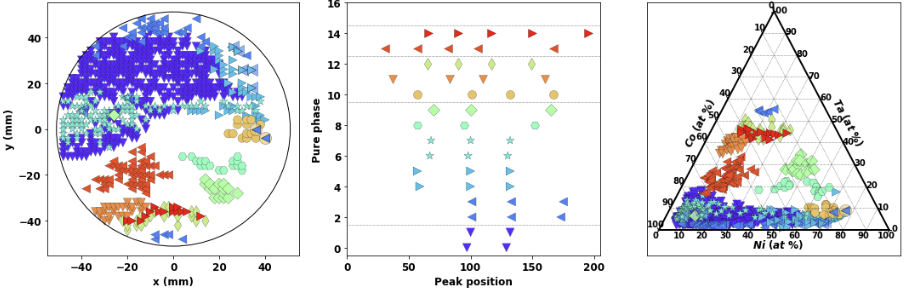}
%}

\caption{Results after using hierarchical clustering to merge similar pure phases from the proposed phase clustering algorithm for Co-Ni-Ta composition space. The proposed algorithm outputs 15 pure phases (Figure~\ref{fig:results}(a)) and then we apply hierarchical clustering on top of the initial pure phases to find the final pure phases (10 in this case). Here, we used a peak adjacency threshold of 2 and threshold count for phase outlier is 15. 
For hierarchical clustering, we used the average difference in peak location in the binary XRD patterns for the pure phase representations as the distance metric.}

\label{fig:hc_res}

\end{figure*}

Since initial pure phases from the proposed phase computation algorithm can contain sets of initial pure phases with similar representations as discussed above, we can either apply hierarchical clustering or combine them manually.
For hierarchical clustering, we used the fclusterdata implementation from hierarchy cluster in SciPy~\cite{scipy} implementations.
We explored several dissimilarity metrics for hierarchical clustering including all of the already defined distance metrics in SciPy~\cite{scipy} and some dissimilarity measures based on peak locations- average difference in peak locations, maximum difference in peak locations, and sum of difference in peak locations.
We observe similar results by using different types of distance metrics.
Figure~\ref{fig:hc_res} illustrates the phase mapping results using hierarchical clustering on the initial pure phases from from the phase computation algorithm using a $th = 1$ and $ot = 12$, for the Co-Ni-Ta composition space. 
We obtained 15 initial pure phases from the phase computation algorithm.
After applying hierarchical clustering, three sets containing two initial pure phases with three peaks, are clustered together.
The efficacy of using hierarchical clustering can be judged by analyzing the pure phase clusters in the ternary plot and circular plot.
For instance, hierarchical clustering clusters together the first two initial pure phases with two peaks.
This might not be correct since those two initial pure phases have different peak locations in binary peak representations (middle subplot) and they are located in different regions in the circular plot.
This suggests that hierarchical clustering may not work perfectly and merge different initial pure phases together, which hints at the inappropriateness of use of hierarchical clustering for XRD analysis even after exhaustive parameter tuning.
Note that the initial pure phases can be effortlessly merged together by domain scientists.
Figure~\ref{fig:hc_res} represent the potential composition-phase diagram output using the proposed approach for Co-Ni-Ta composition space.
Since the initial phase computation outputs for Co-Ti-Ta system in Figure~\ref{fig:results}(c) have distinct initial pure phases, it requires no initial phase merging and represents the potential composition-phase diagram for the system.

\subsection{Domain Expert Verification}

Note that there is no well-defined metric used by domain scientists to quantify the phase clustering results.
We do not use any supervised learning metrics since it would require huge amount of time and manual efforts by domain scientists to label them.
There exist manually computed composition-phase diagrams for both composition spaces used in this study.
The phase diagram for a given material composition system depends on the temperature used during XRD experiments.
This temperature for data collection in our study was in the range of $900^\circ$ C to $1000^\circ$ C.
Nesterenko et al.~\cite{nesterenko1980isothermal} studied the phase equilibria in the Co-Ni-Ta system at $1000^\circ$ C.
Xu et al.~\cite{xu2009phase} investigated the phase equilibria for the Co-Ta-Ti system at $950^\circ$ C.
Comparison between the shape and location of phase regions in the composition phase diagrams using the proposed method and the manually computed phase diagrams by domain scientists shows that 
the final phase diagram for both ternary systems are in good agreement with the manually computed phase diagrams for both systems by domain scientists, having similar phase region shape and location as the computed phase diagram~\cite{nesterenko1980isothermal, xu2009phase}, and concur with the expectations of the domain scientists in our team.
%as shown in Figure~\ref{fig:computed}. 
The proposed approach thus has the potential to drastically reduce the amount of further analysis required for phase indexing by domain scientists.

% \begin{figure*}
% \centering
% \subfigure[Co-Ni-Ta(Source:\cite{nesterenko1980isothermal})]{
% \includegraphics[width = 0.4\linewidth, keepaspectratio=true]{GT_CoNiTa_1000C_Phases.jpg}
% }
% \subfigure[Co-Ti-Ta(Source:\cite{xu2009phase})]{
% \includegraphics[width = 0.4\linewidth, keepaspectratio=true]{GT_CoTiTa_950C_Phases.jpg}
% }

% \caption{Manually Computed Phase Diagrams for the two composition space.}

% \label{fig:computed}

% \end{figure*}

\subsection{Comparison with Current Approaches of Standard Clustering}
\begin{figure*}[!t]
\centering
\subfigure[Co-Ni-Ta using hierarchical clustering]{
\includegraphics[width = 0.45\linewidth, keepaspectratio=true]{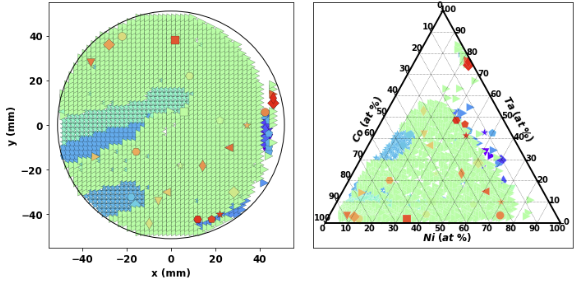}
}
\subfigure[Co-Ti-Ta  using hierarchical clustering]{
\includegraphics[width = 0.45\linewidth, keepaspectratio=true]{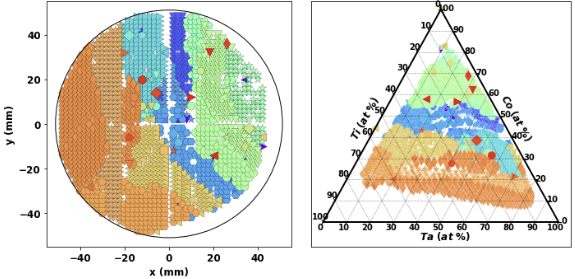}
}
% \caption{Phase diagrams computed using current practice of directly applying hierarchical clustering for the two composition space. Cosine is used as the distance metric in both case, the $hc\_param$ are 0.06103 and 0.48828 for the two cases. There are 53 clusters for the case of Co-Ni-Ta and 65 clusters for the case of Co-Ti-Ta.}
% \label{fig:current}
% \end{figure}

% \begin{figure}[!hbt]
% \centering
\subfigure[Co-Ni-Ta using k-means clustering]{
\includegraphics[width = 0.45\linewidth, keepaspectratio=true]{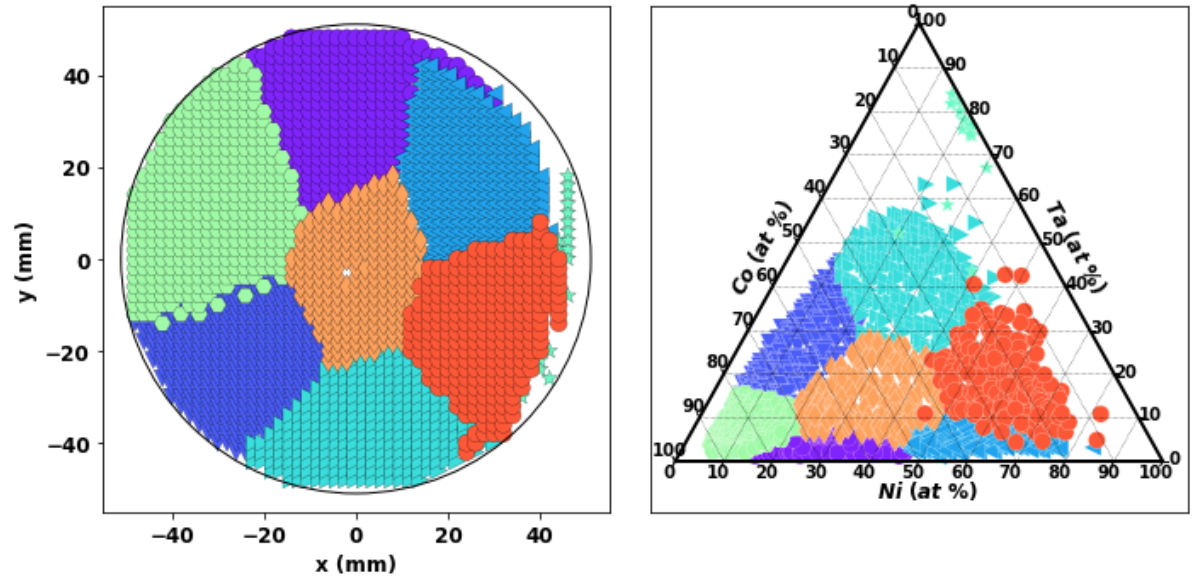}
}
\subfigure[Co-Ti-Ta using k-means clustering]{
\includegraphics[width = 0.45\linewidth, keepaspectratio=true]{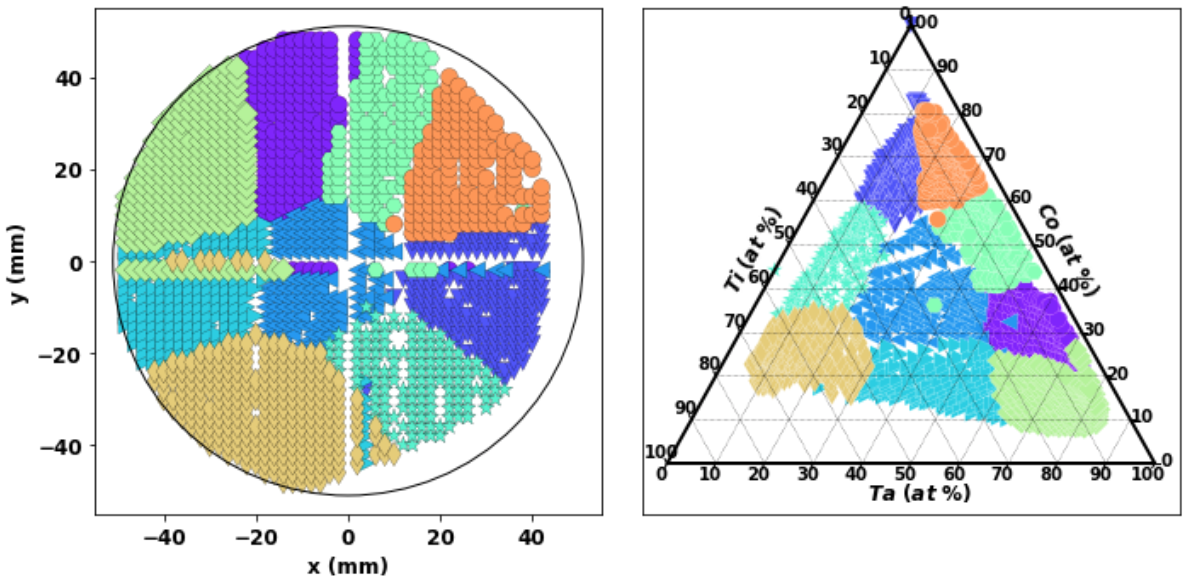}
}
\caption{Phase diagrams computed using current practice of directly applying hard clustering techniques on 1D XRD data. 
For hierarchical clustering, cosine is used as the distance metric in both case, the $hc\_param$ are 0.06103 and 0.48828 for the two cases; there are 53 clusters for the case of Co-Ni-Ta and 65 clusters for the case of Co-Ti-Ta.
For k-means, we used k=9 since there are 9 phases in the final phase diagram verified by domain experts.}
\label{fig:current}
\end{figure*}

Next, we compared our approach against the current practice of directly applying clustering algorithms on the 1D XRD patterns~\cite{bunn2016semi, hattrick2016perspective, iwasaki2017comparison}.
The clustering algorithms used in 1D XRD phase clustering include hierarchical clustering and k-means clustering.
For hierarchical clustering, we experimented using all the existing distance metrics for hierarchical clustering in SciPy~\cite{scipy}, such as 'euclidean', 'cosine', 'seuclidean' and 'correlation'.% 'hamming', 'jaccard', 'chebyshev', 'canberra', 'braycurtis', 'mahalanobis', 'dice', 'kulsinski', 'rogerstanimoto'. 
For each metric, we experimented using 50 values for $hc\_param$ starting at 1000, decreasing by a factor of 2. 
We also experimented using the earth moving distance (EMD) metric and dynmatic time warping (DTW) which are specificially developed for feature preservation with resilience to peak shifting~\cite{iwasaki2017comparison}. For EMD, we started at $hc\_param$ of 1000 and gradually decreasing by a factor of 10.
The experimentation using DTW took infeasible amount of time and did not run to completion.
We used 'distance' as the $hc\_criterion$ and for 'average' as the $method$ after exhaustive parameter search.
Figure~\ref{fig:current} illustrates one of the good clustering results we obtained using hierarchical clustering on the two datasets.
For k-means clustering, we experimented with the value of  $k$ in the range of 5 to 20.

There are a large number of clusters present in the phase clustering results using the current practice of hierarchical clustering; 
Although some clusters are more clear, they represent multiple phase regions combined together.
For Co-Ni-Ta, there are 53 clusters in Figure~\ref{fig:current}(a); the phase regions are overlapping, most of the samples belong to one cluster that is spread throughout the circular and ternary phase diagrams. There is no clear phase region boundaries, and no distinction between pure and mixed phases. 
Similar observations hold for the case of Co-Ti-Ta also, there are 65 clusters in this case.
The phase regions are slightly more clear in the case of Co-Ti-Ta, but the larger phase clusters contain multiple pure phases from ground truth.

If we compare against the phase diagrams manually computed by domain scientists for Co-Ni-Ta~\cite{nesterenko1980isothermal} and Co-Ti-Ta~\cite{xu2009phase}, we find that the phase diagram from the current method hardly match with the phase regions in the ground truth phase regions manually computed by domain scientists; the shapes and location of the phase regions from the current approach of hierarchical clustering are very different from the phase regions in the ground truth.
For phase clustering results with expected number of phases using hierarchical clustering, the phase diagram contains a single phase that spreads throughout the ternary plot containing almost all samples
We observe more clear clusters in the case of k-means clustering,
as shown in Figure~\ref{fig:current}, 
but they hardly resemble with the manually computed composition-phase diagram.
In contrast, the proposed method provides better phase clustering results for both ternary alloys, the phase regions have clear boundaries and the overlapping phase regions can be visualized and corrected/merged if necessary.
This illustrates the benefit of using the proposed phase mapping algorithm over the current practice of using clustering which cannot distinguish between pure phases and mixed phases.
\section{Discussion and Conclusion}
In this paper, we presented a novel algorithm for phase mapping of 1D XRD patterns using a binary peak representation that uses a threshold-based fuzzy dissimilarity measure. 
% The proposed algorithm works by taking into consideration the peak locations and uses a fuzzy peak representation to consider their adjacency which also reduces the amount of computation in further analysis.
The results obtained using current practice of using clustering techniques illustrate that they are not suitable for analysis of 1D XRD patterns. Even though distance metrics which are resilient to peak shifting, such as dynamic time warping (DTW) and earth mover's distance (EMD), can be used with existing clustering algorithms, they do not work for 1D XRD patterns as shown in past works~\cite{iwasaki2017comparison}. On the other hand, we demonstrated that we can compute the phase maps using the proposed threshold-based fuzzy dissimilarity measure that takes into account the adjacency in peak locations to produce a composition-phase diagram that closely resembles the manually computed composition-phase diagram by domain scientists. 
% In addition, the proposed algorithm also provides a discrete binary phase representation for each pure phases which enables simplifies visualization and further analysis by domain scientists. 
The initial pure phase representations enable scientists to make informed decisions regarding the potential composition-phase diagram, and help in determining new pure phases absent in any known phase diagram for a given composition space.
% In the future, we plan to automate the process of parameter selection for clustering in the proposed algorithm and make the software tool available online so that all domain scientists can use it for XRD analysis.
% This illustrates that handling peak adjacency is a key issue in performing phase mapping analysis for 1D XRD patterns.

Although there exist multiple evaluation metrics for clustering analysis, there is no well-defined metric for formally analyzing the phase clustering results other than a qualitative comparison with manually computed composition phase diagram and verification by domain experts. Hence, the results in our study are compared against the manually computed phase diagrams from past literature by domain scientists; the final composition-phase diagrams for both composition spaces using the proposed approach are also verified by domain experts. 

The proposed approach only requires effortless merging of similar initial pure phases if any to obtain the final composition-phase diagram.
The code implementation of the proposed approach is available at \url{https://github.com/NU-CUCIS/IncrementalXRDClustering}.
The datasets used in this study are collected at Argonne National Lab and we do not have proprietary rights to share them openly with public, but are available upon request.
We plan to work towards automating the overall process in the future and release resulting software for domain scientists for the analysis of their high-throughput XRD datasets.
The proposed approach significantly reduces the amount of manual labor in analyzing the large volume of XRD samples coming from high-throughput XRD composition-spread experiments.
The proposed approach takes us closer to achieving complete automation of high-throughput XRD analysis, which is critical to new materials design and discovery, and hence, to the advancement of all scientific fields.

\section{Acknowledgment}

This work was performed under the following financial assistance award 70NANB14H01 and 70NANB19H005 from U.S. Department of Commerce, National Institute of Standards and Technology as part of the Center for Hierarchical Materials Design (CHiMaD), 
DND-CAT located at Sector 5 of the Advanced Photon Source (APS) at Argonne National Lab  supported by DOE under Contract No. DE-AC02-06CH11357,  the MRSEC program of the National Science Foundation (DMR-1720139), and the Soft and Hybrid Nanotechnology Experimental (SHyNE) Resource (NSF NNCI-1542205). Partial support is also acknowledged from DOE awards DE-SC0014330, DE-SC0019358.

\bibliographystyle{unsrt}
\bibliography{bppc}

\end{document}